\documentclass[11pt]{jmlr}
\usepackage{booktabs,times} 
\usepackage{graphicx} 
\usepackage{url}      
\usepackage{amsmath}  
\usepackage[T1]{fontenc}
\usepackage{latexsym,amssymb,enumitem,color,amsmath,booktabs,bm,algorithm,algpseudocode,tikz,graphicx,array,lineno,hyperref,colortbl}
\usepackage{fancyhdr}
\usepackage[table]{xcolor}
\graphicspath{{./Plots/}}
\usetikzlibrary{matrix}
\usetikzlibrary{arrows,plotmarks,decorations.markings,trees,shapes}
\tikzstyle{n}=[ellipse,draw=black!100,fill=black!10,line width=.7pt,minimum width=0.8cm,align=center,text width=1.8cm, text height=.2cm]
\tikzstyle{n2}=[ellipse,draw=black!100,fill=black!0,line width=.7pt,minimum width=0.8cm,align=center,text width=1.5cm, text height=.2cm]
\tikzstyle{nodo}=[ellipse,draw=black!100,fill=black!10,line width=.7pt,minimum width=0.4cm,text width=0.5cm,align=center,minimum height=.5cm]
\tikzstyle{nodo2}=[ellipse,draw=black!100,fill=black!0,line width=.7pt,minimum width=0.6cm,text width=0.6cm,align=center,minimum height=.5cm]
\tikzstyle{arco}=[draw=black!80,line width=1pt, postaction={decorate}, decoration={markings,mark=at position 1.0 with {\arrow[ draw=black!80,line width=.7pt]{>}}}]
\title{Causal Modelling of Support Interventions\\for Student Competency Assessment}
\author{\Name{Francesca Mangili} \Email{francesca.mangili@supsi.ch}\\
\Name{Alessandro Antonucci} \Email{alessandro.antonucci@supsi.ch}\\
  \addr 
  IDSIA - SUPSI\\
Scuola Universitaria Professionale della Svizzera italiana (SUPSI)\\
Istituto Dalle Molle di Studi sull'Intelligenza Artificiale (IDSIA)\\
  Lugano, Switzerland
  \AND
  \Name{Rafael {Caba\~nas de Paz}} \Email{rcabanas@ual.es}\\
  \addr University of Almería\\
  Almería, Spain}
\begin{document}
\thispagestyle{empty}
\pagestyle{fancy}
\fancyhf{}                    
\renewcommand{\headrulewidth}{0pt}
\fancyfoot[C]{\thepage}   
\maketitle
\begin{abstract}
Accurate assessment of students’ competencies is essential for enabling educators to identify individual needs, design targeted interventions, and evaluate the effectiveness of educational strategies. Empirical assessment procedures are typically grounded in psychometric models, such as \emph{item response theory}, which relate students’ competence levels to their performance on assessment tasks. In this paper, we advocate adopting a structural causal modelling approach to educational assessment, moving beyond probabilistic belief updating toward a framework that explicitly supports interventional and counterfactual reasoning. We propose a corresponding protocol for its construction and analyse the practical relevance of forms of reasoning that remain inaccessible to standard associative models, including the explicit modelling of interventions such as hints and the related counterfactual scenario analysis. Although our protocol requires the structural equations to be elicited from experts, the necessary information is purely logical and does not rely on probabilistic, less tenable assumptions. We illustrate the approach using data from an assessment that employs complex tasks designed to measure compulsory school students’ algorithmic skills.
\end{abstract}

\fancypagestyle{firstpage}{%
  \fancyhf{}
  \renewcommand{\headrulewidth}{0pt}
}

\section{Introduction}\label{sec:intro}
Accurate assessment of students' competencies is a cornerstone of effective education. It supports both the design and implementation of large-scale educational strategies by monitoring skill development across the population, and the deployment of personalised interventions, such as those used in tutoring systems, which rely on profiling individual student abilities \citep{godaert2022assessment,koeppen2008current,mikhridinova2024taxonomy}. Developing a formal mathematical model of the competence framework underlying the creation of an assessment instrument can greatly support this goal. \emph{Item response theory} is perhaps the most widely used approach for modelling single competences \citep{koeppen2008current}. Multidimensional extensions have been proposed, but they are not designed to represent mixtures of competencies with individual items. When this is a desideratum, \emph{probabilistic graphical models} (PGMs) are a suitable alternative \citep{culbertson2016bayesian}.

Previous works using PGMs to cope with data collected using batteries of questions led to the adoption of \emph{directed} models, such as \emph{Bayesian networks} (BNs, see, e.g., \citeauthor{vomlel2016}, \citeyear{vomlel2016}) or their more robust, \emph{credal}, i.e., interval-valued, extensions (e.g., \citeauthor{antonucci2021new}, \citeyear{antonucci2021new}). Noisy gates have been later included to ease elicitation by structural assumptions \citep{mangili2022modelling}. The literature has shown that PGMs are excellent tools for mathematically describing a competence model \citep{almond2007modeling,culbertson2016bayesian}. They allow capturing the relationships between multiple competencies and observable behaviours in a flexible and interpretable way, aligning closely with the goals of an assessment. 
Several learner-modelling approaches based on PGMs, such as Bayesian Knowledge Tracing and influence-diagram-based tutoring systems, already incorporate meaningful causal intuitions and, in some cases, explicit modelling of interventions \citep{lin2016intervention} or decision-theoretic components for pedagogical action selection \citep{murray2004looking}. However, these approaches are formulated as probabilistic learner models and do not generally make use of the full Structural Causal Model (SCM) framework, including its explicit treatment of interventions, counterfactual reasoning, and identifiability. This limitation may not substantially affect predictive performance, but it restricts the ability to answer questions that are inherently causal in nature, such as evaluating hypothetical interventions or reasoning about alternative assessment trajectories.

Motivated by these considerations, this paper contributes to evolving such learner modelling protocols from BNs to \emph{structural causal models} (SCMs, \citeauthor{pearl2009causality}, \citeyear{pearl2009causality}). The step looks extremely natural as graphs implementing the BN learner model often have a bipartite structure with arcs from the skills to the questions, where each question receives incoming arcs from the skills that are relevant to answer it (see, e.g., Fig.~\ref{fig:2q}). This aligns with the standard separation between exogenous and endogenous variables in SCMs. Moreover, the gates often used to define the association between skills and answers can be regarded as a noisy version of the \emph{structural equations} (SEs) required for SCM quantification.
Thus, to cope with SCMs, at the elicitation level, we are only required to quantify the structural relations between answers and their parent latent variables that may represent competences, attitudes, question difficulty and chance (e.g., guessing, slipping, over- or under-performing). Uncertainty is represented by probability distributions over these latent variables. Given the observations of the answers, we can back-propagate this information through the SEs and identify the latent distributions compatible with them, thus enabling the computation of robust inferences about the state of competencies for a given group or individual. Importantly, those SCMs also enable explicit modelling of the causal role of task conditions, such as the usage of hints or other forms of support. This is particularly useful when students are allowed to choose the conditions under which they perform a task, since such conditions may influence their response independently of their underlying competence profile.


The paper is organised as follows. We first review the background on PGMs in Sect.~\ref{sec:back}. Our causal elicitation protocol is in Sect.~\ref{sec:protocol}, while the counterfactual inferences of interest are discussed in Sect.~\ref{sec:inference}. Sect.~\ref{sec:case} illustrates the approach in an algorithmic skill assessment based on a battery of \emph{cross-array tasks} (CATs, \citeauthor{piatti2022ct}, \citeyear{piatti2022ct}), whose results are presented in Sect.~\ref{sec:results}. Conclusions and limitations are discussed in Sect.~\ref{sec:conc}. Further details on the inferential complexity of our method are given in App.~\ref{app:comp}, and additional information on the use case is provided in App.~\ref{app:case}.

\section{Background}\label{sec:back}
We use uppercase letters for variables, lowercase letters for states and script letters for the set of possible values, except for the script letter $\mathcal{P}$, which denotes interval-valued probabilities. Thus, a variable $X$ has a generic value $x\in\mathcal{X}$. A \emph{probability mass function} (PMF) over $X$ is denoted as $P(X)$ while a \emph{credal} PMF, that is a set of PMFs, is denoted by $\mathcal{P}(X)$. 

\paragraph{\textbf{Structural Causal Models (SCMs).}}
Given variables $U$ and $V$, a \emph{structural equation} (SE) $f_V^U$ is a map $\mathcal{U}\to\mathcal{V}$, while a \emph{conditional probability table} (CPT) $P(V|U)$ is a collection of PMFs over $V$ indexed by the elements of $\mathcal{U}$. In SCMs, the variables in $\bm{V}$ are called \emph{endogenous}, those in $\bm{U}$ \emph{exogenous}. In this work, endogenous variables --- students' answers and hint usage --- are the observed variables of our model, generated by the latent exogenous variables skill, luck, and help-seeking. A \emph{partially specified} SCM (PSCM) is a collection of SEs $\{ f_V^{\Pi_V}\}_{V\in\bm{V}}$, with $\Pi_V \subset (\bm{V},\bm{U})$, for each $V\in\bm{V}$ \citep{pearl2009causality}. This implicitly defines a directed graph $\mathcal{G}$ whose nodes are in one-to-one correspondence with the variables in $(\bm{U},\bm{V})$ and such that $\Pi_V$ are the \emph{parents} of $V$.
We focus on \emph{semi-Markovian} models, i.e., such that $\mathcal{G}$ is acyclic. A \emph{fully specified} SCM (FSCM) is a PSCM paired with PMFs $\{ P(U) \}_{U\in\bm{U}}$. As a SE defines a CPT made of degenerate, zero/one, PMFs, a FSCM is a special BN defining a joint PMF $P(\bm{V},\bm{U})$ which factorises as $P(\bm{x}) = \prod_{X\in (\bm{V},\bm{U})} P(x|\pi_X)$ with  $(x,\pi_X)$ consistent with $\bm{x}$ \citep{koller2009}. 

\paragraph{\textbf{Causal Inference (in FSCMs).}} 
In BNs, we consider \emph{observational} queries, involving the computation of the posterior PMF for a queried variable, given some evidence about other variables. FSCMs offer more freedom. 
An \emph{intervention} $V=v$ forces the endogenous variable $V$ to take the value $v$ by replacing the original SE with a constant SE, yielding $v$ as output. 
In this setting, the post-interventional PMF for another endogenous variable $W$ is denoted as $P(W_v)$. An interventional query that forces a variable to take values different from the one observed is called here a \emph{counterfactual}. Interventional and counterfactual queries might be computed directly in the FSCM by standard BN inference algorithms. 
Counterfactual queries rely on a \emph{twin} graph, where endogenous variables and their SEs are cloned, while exogenous variables remain shared. This allows simultaneous representation of factual and counterfactual worlds \citep{correa2025counterfactual}.

\paragraph{\textbf{Causal Inference (in PSCMs).}}\label{sec:infpscms}
Since we consider exogenous variables that are not directly observable, in practice, we only cope with a dataset $\mathcal{D}$ of endogenous observations, from which we compute a joint PMF $P(\bm{V})$ but not the exogenous PMFs required by the FSCM definition. Pearl's \emph{do calculus} allows to reduce interventional queries to observational ones by leveraging the causal graph relations \citep{pearl1995causal}. In general the same cannot be done for counterfactual queries. In these cases, back-propagation techniques, such as the causal EM by \citet{zaffalon2024}, allow to find a set of FSCMs compatible with a PSCM. PSCM inference is consequently intended as iterated inference over the compatible FSCMs. 

\section{Educational Tests by PSCMs}\label{sec:protocol}
We present our general protocol to model educational tests by (P)SCMs by starting from the model variables, and then discussing the SE elicitation.

\paragraph{\textbf{Questions and Answers.}} We describe the answers to the questions in the test by a set of endogenous variables $\bm{Q}$, whose actual values are determined through SEs involving other variables as inputs. We want the model to be suitable for a potentially \emph{adaptive} setting, where the sequence of the questions in the test might be changed \citep{antonucci2021new}, and therefore the SEs cannot include answers as inputs.

\paragraph{\textbf{Skills.}} We consider a set of variables $\bm{S}$, each modelling the level at which the test taker masters a relevant competence. We regard these variables as exogenous, thus not directly observable. Yet, unlike the typical case with SCMs, here the exogenous states have a clear semantics implicitly defined by the SEs. Accordingly, skill nodes can be Boolean as well as ordinal or even continuous. Note that a very same skill $S\in\bm{S}$ could be an input of the SEs of two or more questions, thus acting as a \emph{confounder}. 

\paragraph{\textbf{Lucks.}} Besides answer and skill nodes, a set of \emph{luck}\footnote{Here luck plays exactly the role that noise terms play in the structural causal model literature --- akin to a guessing/slip term in classical psychometrics; no further substantive interpretation is intended.}  nodes $\bm{L}$ adding stochasticity to the answers is included. As for the skills, the lucks are exogenous, not directly observable variables. We therefore assume a correspondence between a luck variable and an answer, denoting as $L_Q\in\bm{L}$ the luck associated with $Q\in\bm{Q}$. These variables capture factors such as the difficulty of the question and other contextual elements that may lead students to over- or under-perform with respect to their actual skills.

\paragraph{\textbf{Hints.}} In many testing setups, \emph{hints} specific to the different questions might be available to the students. We describe them by a number of discrete endogenous variables $\bm{H}$. Like for the luck variables, hints are in correspondence with the questions, and we denote as $H_Q$ the hint associated with question $Q\in\bm{Q}$. 

\paragraph{\textbf{Help-Seeking Propensity.}} The decision of a student to request assistance is attributed to an individual \emph{propensity} corresponding to a global variable $R$. It can also be assumed that the skills relevant to question $Q$ also influence the corresponding hint $H_Q$. The actual help  $H_Q$ used for a specific question may still fluctuate around the level determined by the student's general propensity and skill levels. Such deviations, arising either from task characteristics or from chance, are captured by an additional exogenous variable $W_Q$. \\

\begin{figure}[htp!]
\centering
\begin{tikzpicture}
\node[nodo2] (s1)  at (0,0) {\scriptsize $S_1$};
\node[nodo2] (s2)  at (2,0) {\scriptsize $S_2$};
\node[nodo] (h1)  at (-2,0) {\scriptsize $H_1$};
\node[nodo] (h2)  at (4,0) {\scriptsize $H_2$};
\node[nodo2] (l1)  at (-2,-.9) {\scriptsize $L_1$};
\node[nodo2] (l2)  at (4,-.9) {\scriptsize $L_2$};
\node[nodo] (q1)  at (0.,-.9) {\scriptsize $Q_1$};
\node[nodo] (q2)  at (2,-.9) {\scriptsize $Q_2$};
\node[nodo2] (w1)  at (-2,.9) {\scriptsize $W_1$};
\node[nodo2] (w2)  at (4,.9) {\scriptsize $W_2$};
\node[nodo2] (r)  at (2,.9) {\scriptsize $R$};
\draw[arco] (w1) -- (h1);
\draw[arco] (w2) -- (h2);
\draw[arco] (r) -- (h1);
\draw[arco] (r) -- (h2);
\draw[arco] (h1) -- (q1);
\draw[arco] (h2) -- (q2);
\draw[arco] (s1) -- (h1);
\draw[arco] (s1) .. controls (2,.6) .. (h2);
\draw[arco] (s2) -- (h2);
\draw[arco] (s1) -- (q1);
\draw[arco] (s1) -- (q2);
\draw[arco] (s2) -- (q2);
\draw[arco] (l2) -- (q2);
\draw[arco] (l1) -- (q1);
\end{tikzpicture}
\caption{PSCM in Fig.~\ref{fig:2q} augmented with help and propensity nodes.}\label{fig:full}
\end{figure}
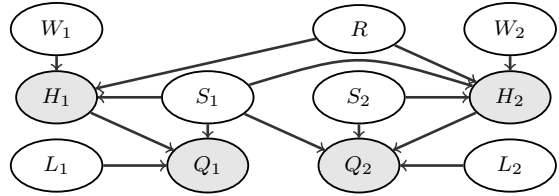

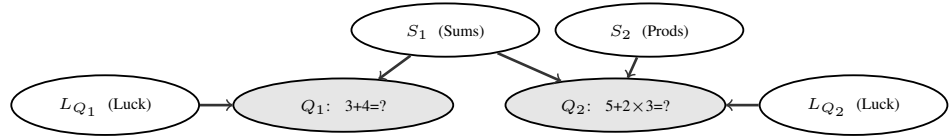
\begin{figure}[htp!]
\centering
\begin{tikzpicture}[scale=0.9]
\node[n2] (s1)  at (3,0) {\tiny $S_1$ (Sums)};
\node[n2] (s2)  at (6,0) {\tiny $S_2$ (Prods)};
\node[n2] (l1)  at (-2,-1.1) {\tiny $L_{Q_1}$ (Luck)};
\node[n2] (l2)  at (9,-1.1) {\tiny $L_{Q_2}$ (Luck)};
\node[n] (q1)  at (1.5,-1.1) {\tiny $Q_1$: 3+4=?};
\node[n] (q2)  at (5.5,-1.1) {\tiny $Q_2$: 5+2$\times$3=?};
\draw[arco] (s1) -- (q1);
\draw[arco] (s1) -- (q2);
\draw[arco] (s2) -- (q2);
\draw[arco] (l2) -- (q2);
\draw[arco] (l1) -- (q1);
\end{tikzpicture}
\caption{Modelling the assessment with two skills and two questions in Ex.~\ref{ex:luck}.\label{fig:2q}}
\end{figure}

Fig.~\ref{fig:full} depicts the PSCM in Fig.~\ref{fig:2q} augmented with the hint and propensity nodes.

\paragraph{\textbf{Structural Knowledge Elicitation.}}
We regard the variables in $\bm{V}:=(\bm{Q},\bm{H})$ as \emph{endogenous}, i.e., directly observable and with actual values induced by SEs, while the variables in $\bm{U}:=(R,\bm{W},\bm{S},\bm{L})$ are parentless \emph{exogenous}, i.e., latent and such that their actual values obey a stochastic process summarised by a marginal PMF for each variable. The SE elicitation requires defining how observable behaviours are generated from latent learner characteristics. Observable variables, such as answers and help requests, are typically defined by the assessment design itself. The main modelling decisions concern the latent variables representing interpretable but unobservable learner characteristics, namely skills and help-seeking propensity, whose states should correspond to meaningfully different observable behaviours, as can be derived from the competence rubrics typically used in educational assessments. For each $Q\in\bm{Q}$, a domain expert is therefore required to elicit the relevant skills $\bm{S}_Q \subseteq \bm{S}$ corresponding to the competencies needed to answer it. In our assumptions, these are also the relevant skills for the hint $H_Q$. As for student attitudes that affect help-seeking behaviours, only propensity is considered here, and it is therefore the only parent of all hint nodes; in principle, however, learner attitudes toward help could be described using more than one exogenous node. The structural equations then formalise how these latent states translate into expected answers and help-seeking patterns, a relationship that defines the operational meaning attributed to the skills and attitudes. By contrast, luck variables $L_Q$ and deviation variables $W_Q$ act as noise terms accounting for slips, guessing, question-specific contextual effects, and other sources of variability not explicitly represented by the learner model. Their state spaces should generally be rich enough to make all observable outcomes reachable through the SE. Ruling out certain outcomes is a legitimate modelling choice, but introduces stronger modelling assumptions and may create incompatibilities with the observed data. Conversely, allowing outcomes that are considered highly unlikely typically results in the learning procedure assigning them very low probabilities whenever they are unsupported by the data. For each $Q\in\bm{Q}$, this yields the SEs $Q = f_Q(\bm{S}_Q,L_Q,H_Q)$ and $H_Q = f_{H_Q}(\bm{S}_Q,R, W_Q)$, giving a complete PSCM specification. It is worth noticing that, under our modelling assumptions, the resulting graph is acyclic by construction, making the model \emph{semi-Markovian} and thereby ensuring more tractable inference \citep{pearl1995causal}.

An example illustrating the protocol (hints excluded for simplicity) is given below.

\begin{example}\label{ex:luck}
Consider a test based on two arithmetic questions: $Q_1$ requires to compute a sum, $Q_2$ both a sum and a product. The answers are described by Boolean variables. Two Boolean skills $S_1$ and $S_2$ describe the additive and multiplicative abilities to be assessed. Answering question $Q_2$ involves both skills, say through a conjunctive relation. Yet, it might also be the case that a student not mastering any of the two skills gives a correct answer because of the corresponding Boolean luck variable $L_{Q_2}$. This can be modelled by the propositional relation $Q_2 = (S_1 \wedge S_2) \vee L_{Q_2}$. For $Q_1$, we have instead a simple disjunction $Q_1=S_1 \vee L_2$. The corresponding graph is depicted in Fig.~\ref{fig:2q}, where we use gray background for nodes associated with manifest, endogenous variables.
\end{example}

\section{Educational Assessments by Causal Inference}\label{sec:inference}
The above protocol for causal modelling of student competency is based on PSCMs. As the inferences required for the assessment are instead based on FSCMs, we might also need the marginal PMFs over the exogenous variables, i.e., $P(R)$, $\{P(S)\}_{S\in\bm{S}}$, $\{P(L)\}_{L\in\bm{L}}$, and $\{P(W)\}_{W\in\bm{W}}$. While these latent PMFs are typically unavailable, we can regard the results of an assessment as a dataset $\mathcal{D}$ of endogenous observations, from which we might learn a PMF $P(\bm{V})$ (e.g., as a BN). As discussed in Sect.~\ref{sec:back}, by the causal EM back-propagation of \cite{zaffalon2024}, a collection of $n$ FSCMs compatible with the given PSCM and the empirical PMF $P(\bm{V})$ can then be derived. Inferences are finally computed separately for each compatible FSCMs. The causal EM procedure is run once per PSCM specification to obtain the (approximated) credal set of compatible FSCMs, not once per individual query; queries are then answered by optimizing the target quantity over that fixed credal set. A \emph{non-identifiable} query $\gamma$ yields a set of different probability values $P_i(\gamma)$ returned by the EM algorithm over the $i$-th compatible FSCM. These are summarised by a, credal, interval-valued probability $\mathcal{P}(\gamma) := [\min_{i=1:n} P_{i}(\gamma),\max_{i=1:n} P_i(\gamma)]$. The inferential complexity of this approach is discussed in App.~\ref{app:comp}.

We can now distinguish between the two types of inferences relevant to educational assessment: \emph{group-level} and \emph{individualised}. Group inferences aggregate information across multiple students, whereas individualised inferences are obtained by further conditioning on the specific observations $(\hat{\bm{q}}, \hat{\bm{h}}) \in \mathcal{D}$ from a given student. Both types of inference are derived from the set of FSCMs compatible with the observations $\mathcal{D}$.

\subsection{Group Inferences}
Group inferences allow assessing the overall skill level of the population under analysis, to evaluate the quality of questions as tools for measuring these skills, and to understand the influence of the help variables on the assessment.

\paragraph{\textbf{Marginal Queries.}}
Unlike most common approaches to SCMs, our models assign definite semantics to the states of the exogenous variables by implicitly defining them through the SEs. This makes the marginal queries on these variables interpretable. As these variables correspond to parentless nodes, their marginal PMFs are already available in the $n$ FSCMs we compute by the causal EM. For each skill $S\in\bm{S}$, the interval-valued marginal PMFs $\mathcal{P}(S)$ summarises the relative distribution in the population of interest, thus describing the group characteristics, specifically their proficiency and attitude. Additional information about propensity is provided by the marginal $\mathcal{P}(R)$. It is also important to look at the marginal luck information $\mathcal{P}(L_Q)$, which measures how well a question aligns with the skill level it is expected to assess. An informative question $Q$ should have a high probability that the associated luck node $L_Q$ is in a neutral state (neither good nor bad luck). If the probability of experiencing luck (good or bad) on a question is high, students will likely perform better or worse on that question than expected given their abilities, e.g., because the question is simpler or more difficult than it was meant to be. 

\paragraph{\textbf{Necessity and Sufficiency.}} Evaluating the causal effect of the hint $H_Q$ towards the answer to $Q$ can be naturally achieved by Pearl's counterfactual probabilities of \emph{necessity} ($\mathrm{PN}_Q$) and \emph{sufficiency} ($\mathrm{PS}_Q$)
\citep{pearl1999probabilities}. The former quantifies how necessary the help is for a correct answer, i.e., if both variables are Boolean, at the group level, we have $\mathrm{PN}_{Q}:=P(Q_{H_Q=0}=0|H_Q=1, Q=1)$, i.e., the probability that the student would have answered incorrectly had they not received help (formally, this corresponds to the event $Q_{H_Q=0}=0$), given that they did receive help ($H_Q=1$) and answered correctly ($Q=1$). Similarly, $\mathrm{PS}_{Q}:=P(Q_{H_Q=1}=1|H_Q=0, Q=0)$ measures the probability that the student would have answered correctly if help had been provided, given that they did not receive help and answered incorrectly. 

In applications (e.g., see Sect.~\ref{sec:case}), we often cope with \emph{ordinal}, non-binary variables. If $q_1 < q_2 < \ldots$ and $h_1 < h_2 < \ldots$ are the orders of the states of $Q$ and $H_Q$, we generalise $\mathrm{PN}_Q$ as: $\mathrm{PN}_{Q}(q_j,h_k):=P(Q_{H_Q=h_{k-1}} < q_j|H_Q=h_k,Q \geq q_j)$, that is the probability that the student would perform at level $Q < q_j$ if given help $H_Q=h_{k-1}$ knowing that with higher help ($H_Q= h_{k}$) they performed better. Similarly, we define $\mathrm{PS}_{Q}(q_j,h_k):=P(Q_{H_Q=h_{k}}\geq q_j|H_Q =h_{k-1}, Q < q_j)$, that is the probability that the student would perform at level $Q\geq q_j$ if given help $H_Q\geq h_k$ knowing that with less help ($H = h_{k-1}$) they performed worst.
We consider analogous counterfactual quantities,  i.e., $\mathrm{PN}_{S}(q_j,s_k)$ and $\mathrm{PS}_{S}(q_j,s_k)$ for each $S\in\bm{S}$, to decide how informative a specific question about the skills is. 

Queries such as $P(Q|H_Q)$ or $P(Q|S)$ also describe the relationship between the question and help or question and skill variables. However, since they are purely observational, they may be influenced by confounding factors and cannot isolate the causal effect of help on performance. E.g., students might perform better without help than with it, simply because more proficient students tend not to request help, while those who do are more likely to fail.

\paragraph{\textbf{Computation of Generalised PN and PS.}} As in their standard formulations, the generalised versions of PN and PS can be computed by first constructing the twin network \citep{balke1994counterfactual,cabanas2025bayesian}, which is an SCM containing endogenous variables for both the real and hypothetical scenarios.  This is obtained by duplicating the sub-graph composed of the endogenous nodes in the real scenario and then applying the intervention.  In the twin network, the endogenous nodes in both scenarios share the same exogenous parents, except for the intervened variables. In the generalised setting, however, exogenous variables with evidence (such as the node $S$ when computing its $\mathrm{PS}_S$ and $\mathrm{PN}_S$) are also duplicated and do not connect the two sub-graphs. Furthermore, for the conditioned variable, the probability is computed over a subset of states rather than a single state.

\subsection{Personalised Inferences}
Individualised inferences enable the analysis of a single student's proficiency by examining their specific answers $\hat{\bm{q}}$ in light of any assistance $\hat{\bm{h}}$ used.

\paragraph{\textbf{Student Profile.}} The posterior PMF ${P}(U|\hat{\bm{q}},\hat{\bm{h}})$, where $U$ could be a skill $S$ or the propensity $R$, describes the student’s profile as inferred from the assessment. These distributions reflect how the observed responses inform us about each student's underlying abilities and behavioural tendencies based on the provided PSCM.

\paragraph{\textbf{Individual Counterfactuals.}} In most situated assessment settings, it is not feasible to observe how each student would perform under all possible help conditions for a given task. For example, when evaluating the ability to solve a specific expression $Q$, we may only see the student attempting it mentally; if they fail, we cannot directly infer whether they would have succeeded with paper and pencil.
Counterfactual inferences can enhance assessment in the presence of variable help-seeking behaviours by estimating, for example, the counterfactual probability that a student would have solved the equation using paper given that they did not ask for help and failed: ${P}(Q_{H_Q=1}=1|Q=0,H_Q=0)$,
Since we are interested in characterising a specific student, such inference must be conditioned on their full set of observations, i.e., ${P}(Q_{H_Q=1}=1|Q=0,H_Q=0,\hat{\bm{q}}_{-Q},\hat{\bm{h}}_{-Q})$, where $\hat{\bm{q}}_{-Q}$ denote the set of all the answers in $\hat{\bm{q}}$ apart from $Q$.
To analyse how a student performance would change under different levels of help, we consider the counterfactuals ${P}(Q_{H_Q=h_k}=q_j|\hat{\bm{q}},\hat{\bm{h}})$ which express the probability that a student would have answered at level $q_j$ had they received help $h_k$ given their observed levels $(\hat{\bm{q}},\hat{\bm{h}})$. Such counterfactuals are informative only when the hypothetical levels $h_k$ and $q_j$ change in a direction that is compatible with the expected effect of help. Conversely, they are not meaningful in scenarios where a higher‑level answer would result from a lower-level help, or vice versa.



\section{Use Case}\label{sec:case}
We showcase the proposed approach on a use case about the assessment of algorithmic skills of compulsory school pupils based on the CAT battery introduced by \citet{piatti2022ct} and modelled by \cite{mangili2022modelling} as a BN based on the data collected from 109 students. More details about the assessment protocol are given in App.~\ref{app:case}.

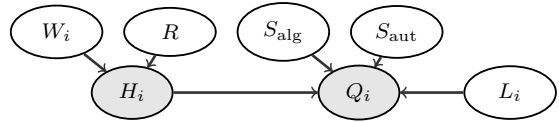
\begin{figure}[htp!]
\centering
\begin{tikzpicture}
\node[nodo2] (s0)  at (1,0) {\scriptsize $S_{\mathrm{alg}}$};
\node[nodo2] (s1)  at (2.5,0) {\scriptsize $S_{\mathrm{aut}}$};
\node[nodo] (h1)  at (-1,-.8) {\scriptsize $H_i$};
\node[nodo2] (l1)  at (4,-.8) {\scriptsize $L_i$};
\node[nodo] (q1)  at (2,-.8) {\scriptsize $Q_i$};
\node[nodo2] (w1)  at (-2,0) {\scriptsize $W_i$};
\node[nodo2] (r)  at (-0.5,0) {\scriptsize $R$};
\draw[arco] (w1) -- (h1);
\draw[arco] (r) -- (h1);
\draw[arco] (h1) -- (q1);
\draw[arco] (s0) -- (q1);
\draw[arco] (s1) -- (q1);
\draw[arco] (l1) -- (q1);
\end{tikzpicture}
\caption{A slice of the causal network for the CAT assessment.}
\label{fig:CATmodel}
\end{figure}

\paragraph{\textbf{Causal CAT Model.}}
Twelve question nodes $\{Q_i\}_{i=1}^{12}$ taking values in $\mathcal{Q} = \{\mathrm{fail},\mathrm{0D},$ $\mathrm{1D},\mathrm{2D}\}$ describe the algorithm used to solve each CAT scheme. For each $Q$, we introduce a \emph{hint} $H_Q$ taking values in $\mathcal{H} = \{\mathrm{none},\mathrm{scheme},\mathrm{feedback}\}$, a \emph{luck} $L_Q$ taking values in $\mathcal{L} = \{\mathrm{very}\,\mathrm{bad},\mathrm{bad},\mathrm{neutral},\mathrm{good},\mathrm{very}\,\mathrm{good}\}$, and $W_Q$ taking values in $\mathcal{W}= \{-1,0,1\}$. 
Following the assumptions of \citet{piatti2022ct}, we define two skills: $S_{\mathrm{alg}}$ describing the \emph{algorithmic competence} and taking values in $\mathcal{S}_{\mathrm{alg}}=\{\mathrm{0D},\mathrm{1D},\mathrm{2D}\}$, and $S_{\mathrm{aut}}$ describing \emph{autonomy} and taking values in $\mathcal{S}_{\mathrm{aut}} = \{\mathrm{feedback},\mathrm{scheme},\mathrm{none}\}$. These two skills are parents of all $Q$ variables, while, to ease elicitation and to reduce the model’s structural and computational complexity, they do not directly influence the hint variables. The hint variables have the propensity $R$ as a parent, taking values in $\mathcal{R}=\{\mathrm{feedback},\mathrm{scheme},\mathrm{none}\}$. All variables are ordinal and their states are presented in ascending order. The common structure shared by all answer and help nodes is in Fig.~\ref{fig:CATmodel}. Each answer $Q_i$ has four parents: skills $S_{\mathrm{alg}}$ and $S_{\mathrm{aut}}$ and the question-specific help and luck, $H_i$ and $L_i$. Each help has two parents: the student-specific \emph{propensity} $R$ and the question-specific node $W_i$.


The SE of $Q_i$ is specified by means on the following rules. (i) A skill level $S_{\mathrm{alg}} = k$ allows applying algorithm $k$, provided that the \emph{help} node state matches the student autonomy, i.e., $H_i = S_{\mathrm{aut}}$. (ii) If $H_i$ is greater or smaller than $S_{aut}$, an algorithm of equally higher or lower complexity is applied. (iii) Then, the level of $Q_i$ is shifted according to the value of the luck $L_i$: $L_Q = $ neutral leaves the level of $Q_i$ unchanged; bad/good luck decreases/increases it by one level; very bad/very good luck decreases/increases the complexity by two levels. (iv) Finally, all shifts of $Q_i$ are clipped to its allowed range, from  $\mathrm{fail}$ to $\mathrm{2D}$. To represent these rules by a compact formal way, we denote as $\rho_V$  the position of $v$ within the ordered set of possible values of $V$. For instance, since the ordered values of $H$ are \{feedback, scheme, none\}, then  $\rho_H=0$ if $H = \mathrm{feedback}$ whereas $\rho_H=2$ if $H=\mathrm{none}$. The above rules can then be summarised by $\rho_{Q_i} = \min\left(3, \max \left( 0, \rho^*\right)\right)$ where:
\[
\begin{aligned}
\rho^* &= \left(\rho_{S_{\mathrm{alg}}}+1\right) + \left(\rho_{H_i}-\rho_{S_{\mathrm{aut}}}\right)+(\rho_{L_i}-2)\,.
\end{aligned}
\label{eq:SE_Q}
\]


The SEs for the hint variables assume that, in the absence of question-specific deviations encoded by $W_i$, the value of $H_i$ coincides with the individual propensity for hints $R$. Otherwise, $H_i$ is adjusted by the deviation term $W_i$, either increasing or decreasing its value, and the result is clipped to the admissible range for $H_i$, which goes from $\mathrm{feedback}$ to $\mathrm{none}$. Formally, this is expressed as: 
\[\rho_{H_i} = \min\left(2, \max\left(0, \rho_R + W_i\right)\right).
\]



\section{Results}\label{sec:results}
In this section, we illustrate how group and individual inferences can be used to analyse an assessment instrument and a student's profile. All inferences were computed using  
the \emph{bcause}\footnote{\href{https://github.com/PGM-Lab/bcause}{github.com/PGM-Lab/bcause}.} open-source software. 
Before illustrating such inferences, we evaluate the expert-elicited FSCM model against a baseline consisting of a BN learned directly from data (both structure and parameters). For each approach, we consider a five-fold cross-validation scheme and compute the log-likelihood on the held-out test set, as well as the posterior probability of each answer $Q_i$ and hint $H_i$ given the observations for all other questions $(Q_j, H_j)_{j=1:12, j\neq i}$. Predictions for $Q_i$ and $H_i$ were then obtained by selecting the most probable state according to these posterior distributions.  Finally, we computed the mean and standard deviation of the average log-likelihood and of the prediction accuracy for questions and hints over the 5 repetitions.  
The results show a test log‑likelihood of -277$\pm$19.1 (standard deviation) for the BN and -287$\pm$18.5 for the FSCM, and a prediction accuracy of 0.84$\pm$0.02 (answers) and 0.84$\pm$0.04 (hints) for the BN, compared to 0.77$\pm$0.02 (answers) and 0.82$\pm$0.04 (hints) for the FSCM. These findings indicate that the modelling assumptions introduced in our framework are compatible with the observed data, although they lead to a reduction in predictive accuracy compared to a purely data‑driven approach. Nevertheless, an expert‑elicited causal model offers advantages that go beyond predictive performance: it is more interpretable and supports explicit causal reasoning. 

\paragraph{\textbf{Group Inferences.}}
For the marginal queries on $S_{\mathrm{alg}}$ we have $P(S_{\mathrm{alg}}=\mathrm{0D})\in [0.02,0.08]$, $P(S_{\mathrm{alg}}=\mathrm{1D})\in[0.02,0.98]$, and $P(S_{\mathrm{alg}}=\mathrm{2D}) \in [0,0.90]$. These results are almost vacuous about level 1D and 2D, indicating that any prior distribution over these group skill levels is compatible with the observations, but shows the very low probability of level 0D. Conversely, for the propensity, we have sharp values (i.e., the query is identifiable): $P(R=\mathrm{feedback})=0.02$ and $P(R=\mathrm{none})=0.37$ show that while only few students are inclined to ask for feedback, a relevant part of them prefers to avoid external support. For $S_{\mathrm{aut}}$, we observe $P(S_{\mathrm{aut}}=\mathrm{feedback})\in [0.92,1]$, $P(S_{\mathrm{aut}}=\mathrm{scheme})\in[0,0]$, and $P(S_{\mathrm{aut}}=\mathrm{none}) \in [0,0.08]$. These results suggest that the lowest autonomy level is highly likely, which is consistent with the young age of the students, while the intermediate scheme level is essentially impossible.

The marginal probabilities of the luck variable in Tab.~\ref{tab:luck} can provide a diagnostic insight into the informativeness of different questions. Specifically, the first six questions appear to be more informative than $Q_7$, $Q_8$ and $Q_9$, since the probability of experiencing very good luck in the former never exceeds 15\% while it can be above 30\% in the latter. This indicates that students may largely over-perform in these questions, suggesting that they are not well calibrated. 
Interestingly, it is very unlikely to experience good or very good luck in $Q_{12}$, while the probability of bad luck can be as high as 0.76. This suggests that the scheme is particularly challenging, making underperformance more likely than over-performance.

\begin{table}[htp!]
\centering
\scriptsize
\begin{tabular}{lp{1mm}ccp{3mm}ccp{3mm}ccp{3mm}ccp{3mm}cc}
\hline
$L=$&&\multicolumn{2}{c}{very bad}&&\multicolumn{2}{c}{bad}
&&\multicolumn{2}{c}{neutral}&&\multicolumn{2}{c}{good}&&\multicolumn{2}{c}{very good}\\
\hline
$P(L_1)$&&
\cellcolor{blue!10}0.00&\cellcolor{orange!10}0.00&&
\cellcolor{blue!10}0.01&\cellcolor{orange!10}0.02&&
\cellcolor{blue!10}0.02&\cellcolor{orange!90}0.88&&
\cellcolor{blue!10}0.08&\cellcolor{orange!90}0.93&&
\cellcolor{blue!10}0.00&\cellcolor{orange!10}0.04\\
$P(L_2)$&&
\cellcolor{blue!10}0.00&\cellcolor{orange!10}0.01&&
\cellcolor{blue!10}0.01&\cellcolor{orange!10}0.02&&
\cellcolor{blue!10}0.02&\cellcolor{orange!90}0.92&&
\cellcolor{blue!10}0.04&\cellcolor{orange!90}0.94&&
\cellcolor{blue!10}0.00&\cellcolor{orange!10}0.03\\
$P(L_3)$&&
\cellcolor{blue!10}0.00&\cellcolor{orange!10}0.00&&
\cellcolor{blue!10}0.00&\cellcolor{orange!10}0.03&&
\cellcolor{blue!10}0.04&\cellcolor{orange!90}0.81&&
\cellcolor{blue!10}0.08&\cellcolor{orange!90}0.82&&
\cellcolor{blue!10}0.00&\cellcolor{orange!20}0.15\\
$P(L_4)$&&
\cellcolor{blue!10}0.00&\cellcolor{orange!10}0.00&&
\cellcolor{blue!10}0.04&\cellcolor{orange!10}0.02&&
\cellcolor{blue!10}0.00&\cellcolor{orange!90}0.98&&
\cellcolor{blue!10}0.00&\cellcolor{orange!90}0.96&&
\cellcolor{blue!10}0.00&\cellcolor{orange!10}0.01\\
$P(L_5)$&&
\cellcolor{blue!10}0.00&\cellcolor{orange!10}0.00&&
\cellcolor{blue!10}0.00&\cellcolor{orange!10}0.00&&
\cellcolor{blue!10}0.00&\cellcolor{orange!90}0.97&&
\cellcolor{blue!10}0.02&\cellcolor{orange!90}0.99&&
\cellcolor{blue!10}0.00&\cellcolor{orange!10}0.01\\
$P(L_6)$&&
\cellcolor{blue!10}0.00&\cellcolor{orange!10}0.01&&
\cellcolor{blue!10}0.01&\cellcolor{orange!10}0.04&&
\cellcolor{blue!10}0.04&\cellcolor{orange!90}0.95&&
\cellcolor{blue!10}0.01&\cellcolor{orange!90}0.95&&
\cellcolor{blue!10}0.00&\cellcolor{orange!10}0.01\\
$P(L_7)$&&
\cellcolor{blue!10}0.00&\cellcolor{orange!10}0.00&&
\cellcolor{blue!10}0.02&\cellcolor{orange!50}0.52&&
\cellcolor{blue!20}0.15&\cellcolor{orange!50}0.51&&
\cellcolor{blue!10}0.00&\cellcolor{orange!30}0.33&&
\cellcolor{blue!10}0.00&\cellcolor{orange!30}0.33\\
$P(L_8)$&&
\cellcolor{blue!10}0.02&\cellcolor{orange!10}0.03&&
\cellcolor{blue!10}0.02&\cellcolor{orange!50}0.53&&
\cellcolor{blue!10}0.08&\cellcolor{orange!50}0.51&&
\cellcolor{blue!10}0.00&\cellcolor{orange!30}0.33
&&\cellcolor{blue!10}0.00&\cellcolor{orange!30}0.33\\
$P(L_9$)&&
\cellcolor{blue!10}0.03&\cellcolor{orange!11}0.11&&
\cellcolor{blue!10}0.09&\cellcolor{orange!50}0.44&&
\cellcolor{blue!10}0.00&\cellcolor{orange!40}0.41&&
\cellcolor{blue!10}0.00&\cellcolor{orange!10}0.00&&
\cellcolor{blue!60}0.47&\cellcolor{orange!50}0.47\\
$P(L_{10})$&&
\cellcolor{blue!10}0.04&\cellcolor{orange!10}0.05&&
\cellcolor{blue!10}0.01&\cellcolor{orange!58}0.58&&
\cellcolor{blue!20}0.19&\cellcolor{orange!50}0.53&&
\cellcolor{blue!10}0.16&\cellcolor{orange!50}0.51&&
\cellcolor{blue!10}0.00&\cellcolor{orange!30}0.25\\
$P(L_{11})$&&
\cellcolor{blue!10}0.04&\cellcolor{orange!10}0.05&&
\cellcolor{blue!10}0.01&\cellcolor{orange!20}0.20&&
\cellcolor{blue!20}0.20&\cellcolor{orange!50}0.53&&
\cellcolor{blue!10}0.16&\cellcolor{orange!50}0.51&&
\cellcolor{blue!10}0.00&\cellcolor{orange!25}0.24\\
$P(L_{12})$&&
\cellcolor{blue!10}0.05&\cellcolor{orange!10}0.08&&
\cellcolor{blue!10}0.05&\cellcolor{orange!75}0.76&&
\cellcolor{blue!10}0.17&\cellcolor{orange!70}0.73&&
\cellcolor{blue!10}0.00&\cellcolor{orange!20}0.17&&
\cellcolor{blue!10}0.00&\cellcolor{orange!10}0.00\\
\hline
\end{tabular}
\caption{Lower and upper bounds of the marginal probabilities of the luck variables.}
\label{tab:luck} 
\end{table}

As an example of PN and PS probabilities, Tab.~\ref{tab:gPNS} shows the bounds of the probabilities of necessity and sufficiency for the help and the \emph{algorithmic} skill associated with $Q_6$. 
These can sometimes result in rather wide intervals, making the inferences nearly vacuous. For instance, the probability of necessity of $S_{alg}=\mathrm{1D}$ for $Q_6=\mathrm{1D}$ could take any value between 0.11 and 0.95.  
Most inferences, however, are quite informative. E.g., the lower bound of the probability of sufficiency of $H_6 = \mathrm{feedback}$ for $Q_6=\mathrm{0D}$ is 0.94, meaning that providing a blank cross-array and visual feedback is very likely to enable a student to solve the task with at least a 0D algorithm, regardless of their skill level. Similarly, the lower bound of the probability that $S_{\mathrm{alg}}=\mathrm{1D}$ is necessary for $Q_6=\mathrm{1D}$ is 0.68 and for $Q_6=\mathrm{2D}$ is 0.96, and the lower bound of $PN(q_6=2D,S_{\mathrm{alg}}=\mathrm{2D})$ is 0.68. This suggests that a student with only level $S_{\mathrm{alg}} = \mathrm{0D}$ would fail delivering an algorithm above 0D complexity in this task and one with $S_{\mathrm{alg}}=1D$ is unlikely to deliver 2D complexity.  We also observe that skill levels 1D and 2D are both likely to be sufficient for solving the question with a 0D algorithm, and that 2D-level skill is also likely sufficient for solving it with a 1D algorithm. On the other hand, the probability that skill level 1D is sufficient for applying a 2D algorithm is at most 0.06. Together, these results indicate that this question may be well suited to discriminate the level of the  $S_{\mathrm{alg}}$ skill that students possess.

\begin{table}[htp!]
\centering
\scriptsize
\begin{tabular}{lp{1mm}ccp{1mm}ccp{1mm}ccp{1mm}cc}
\hline
&&
\multicolumn{2}{c}{$q_6=$0D}&&
\multicolumn{2}{c}{$q_6=$1D}&&
\multicolumn{2}{c}{$q_6=$2D}\\
\hline
$\mathrm{PN}(q_6,h_6=\mathrm{scheme})$&&
\cellcolor{blue!10}0.00&\cellcolor{orange!10}0.02&&
\cellcolor{blue!10}0.02&\cellcolor{orange!10}0.02&&
\cellcolor{blue!60}0.73&\cellcolor{orange!100}1.00\\
$\mathrm{PN}(q_6, h_6=\mathrm{feedback})$&&
\cellcolor{blue!10}0.00&\cellcolor{orange!10}0.01&&
\cellcolor{blue!10}0.01&\cellcolor{orange!10}0.01&&
\cellcolor{blue!57}0.57&\cellcolor{orange!100}1.00\\
$\mathrm{PS}(q_6,h_6=\mathrm{scheme})$&&
\cellcolor{blue!10}0.00&\cellcolor{orange!10}0.00&&
\cellcolor{blue!10}0.00&\cellcolor{orange!10}0.03&&
\cellcolor{blue!10}0.00&\cellcolor{orange!10}0.00\\
$\mathrm{PS}(q_6, h_6=\mathrm{feedback})$&&
\cellcolor{blue!94}0.94&\cellcolor{orange!97}0.97&&
\cellcolor{blue!30}0.29&\cellcolor{orange!80}0.80&&
\cellcolor{blue!80}0.80&\cellcolor{orange!91}0.91\\
\hline
$\mathrm{PN}(q_6, S_{alg} = 1D)$&&
\cellcolor{blue!10}0.04&\cellcolor{orange!90}0.85&&
\cellcolor{blue!70}0.68&\cellcolor{orange!90}0.93&&
\cellcolor{blue!90}0.96&\cellcolor{orange!90}1.00\\
$\mathrm{PN}(q_6,S_{alg} = 2D)$&&
\cellcolor{blue!10}0.01&\cellcolor{orange!10}0.03&&
\cellcolor{blue!10}0.04&\cellcolor{orange!90}0.85&&
\cellcolor{blue!70}0.68&\cellcolor{orange!90}0.93\\
$\mathrm{PS}(q_6,S_{alg} = 1D)$&&
\cellcolor{blue!80}0.80&\cellcolor{orange!90}0.96&&
\cellcolor{blue!10}0.11&\cellcolor{orange!90}0.95&&
\cellcolor{blue!10}0.04&\cellcolor{orange!10}0.06\\
$\mathrm{PS}(q_6, S_{alg} =2D)$&&
\cellcolor{blue!80}0.76&\cellcolor{orange!90}1.00&&
\cellcolor{blue!80}0.80&\cellcolor{orange!90}0.96&&
\cellcolor{blue!10}0.11&\cellcolor{orange!90}0.95\\
\hline
\end{tabular}
\caption{Lower and upper bounds of PN and PS for help and \emph{algorithmic} skill.}
\label{tab:gPNS}
\end{table}

Moreover, some patterns align with our expectations. The PN of both the help and \emph{algorithmic} skill tends to increase with the level of $Q$, suggesting that the probability of needing assistance or a higher skill level becomes more pronounced when attempting to communicate more complex algorithms. Conversely, their sufficiency may decrease, indicating that the probability of successfully applying a more complex algorithm solely based on increased assistance or skill level diminishes as algorithmic complexity increases.

\paragraph{\textbf{Individual Inferences.}}
To illustrate individual-level inferences, we first focus on two extreme cases: a student failing all tasks despite having both feedback and scheme available, and, vice versa, a student who always uses a 2D algorithm without any help. We summarise the competence profiles inferred from their responses by the posterior marginal probabilities of the exogenous student-level variables: $S_{\mathrm{alg}}$, $S_{\mathrm{aut}}$, and $R$ given the observations $(\hat{\mathbf{q}}, \hat{\mathbf{h}})$ collected for that student. For notational simplicity, we will omit the conditioning on $(\hat{\mathbf{q}}, \hat{\mathbf{h}})$ in what follows; all reported probabilities should nevertheless be interpreted as posterior probabilities. In the first case, we obtain identifiable posterior probabilities equal to 1 for the lower levels of \(S_{\mathrm{alg}}\), \(S_{\mathrm{aut}}\), and \(R\). In the second case, instead, only \(P(R=\mathrm{none})=1\), while for \(S_{\mathrm{alg}}\) we obtain \(P(S_{\mathrm{alg}}=\mathrm{0D})=0\), \(P(S_{\mathrm{alg}}=\mathrm{1D}) \in [0,0.40]\), and \(P(S_{\mathrm{alg}}=\mathrm{2D}) \in [0.60,1]\). For \(S_{\mathrm{aut}}\), we obtain \(P(S_{\mathrm{aut}}=\mathrm{feedback}) \in [0,0.60]\), \(P(S_{\mathrm{aut}}=\mathrm{scheme})=0\), and \(P(S_{\mathrm{aut}}=\mathrm{none}) \in [0.4,1]\). These results highlight a limitation of the question battery: it cannot conclude with high confidence that a student possesses the highest skill level even when all tasks are completed at the maximum level. For the algorithmic skill, this is probably due to the excessive influence of the luck variable; for autonomy, instead, it might stem from the high prior assigned to the lowest autonomy level in this specific population. 

Then, we focus on a student who solved $Q_8$ and $Q_{12}$ using a 1D algorithm and all other tasks using a 2D. The student requested a blank cross-array with $Q_5$, $Q_7$, $Q_9$, $Q_{10}$, $Q_{12}$, and feedback for $Q_2$ and $Q_8$.
The posteriors are \(P(R=\mathrm{scheme})\!=\!1\), \(P(S_{\mathrm{alg}}\!=\!\mathrm{0D})\!=\!0\), \(P(S_{\mathrm{alg}}\!=\!\mathrm{1D})\!\in\![0,0.76]\), \(P(S_{\mathrm{alg}}\!=\!\mathrm{2D})\!\in\![0.24 ,1]\), \(P(S_{\mathrm{aut}}\!\!=\!\mathrm{feedback})\!\in\! [0,0.24]\), \(P(S_{\mathrm{aut}}\!\!=\!\mathrm{schema})\!=\!0\), and \(P(S_{\mathrm{aut}}\!\!=\!\mathrm{none})\!\in\![0.76,1]\).
As examples of counterfactuals, we consider $Q_5$ and $Q_{12}$ where the student asked for the scheme help. We investigate whether the student would have solved $Q_5$ at the 2D level even without this support. Additionally, as $Q_{12}$ was solved using a 1D algorithm, we ask whether the additional support of feedback could have enabled the application of a 2D algorithm. These questions are addressed through counterfactuals: ${P}(Q_{5,H_5\!=\!\text{none}}| Q_5=2D,H_5\!=\!\text{scheme})$ and ${P}(Q_{12,H_{12}\!=\!\mathrm{feedback}} |Q_{12}\!=\!1D, H_{12}\!=\!\mathrm{scheme})$. The former yields probability one for $Q_{5}=\mathrm{2D}$, i.e., we are certain that the student would have been able to use a 2D algorithm even without help. 
 For $Q_{12}$, the counterfactual probabilities are $[0, 0]$, $[0.24, 1]$ and $[0, 0.76]$ for algorithm complexities 0D, 1D and 2D. This indicates uncertainty regarding whether 1D or 2D would have been used with more assistance. Specifically, the 1D probability is at least 0.24 and at most 0.76 for 2D.

\section{Limitations and Conclusions}\label{sec:conc}
We explored the application of SCMs with SEs specified by domain experts and informed by student assessment data, enabling the construction of a learner model that explicitly represents causal relationships and supports interventional and counterfactual reasoning. Through an illustrative use case, we showed how such a structure can be used to address key assessment-related objectives, including the evaluation of assessment instruments, the analysis of intervention effects, and learner profiling. Our analysis highlights the potential of this framework, particularly in settings where students can influence the conditions under which they complete tasks: here, counterfactual reasoning offers a principled way to disentangle behavioural factors --- such as help-seeking --- from underlying proficiency. These properties suggest possible applications in adaptive assessment scenarios, where decisions about offering support could be informed by interventional queries. 
While not directly comparable to knowledge-tracing models, since differently from the single-shot assessment considered here, these primarily target the temporal dynamics of learning across repeated attempts, we view our approach as complementary to this line of research, and future work may investigate its extension to dynamic causal learner-modelling frameworks, enabling causal reasoning within temporally grounded models of learning.

At the same time, the present work should be primarily seen as a methodological exploration, and several limitations follow from this positioning. While the learner model is instantiated using data from a real assessment, its role here is illustrative: structure and parameterisation are not really validated, having been tested only on a small dataset, against a BN learned directly from data, and only for predictive accuracy. Benchmarking is limited by the lack of suitable empirically validated competence models accompanied by large interventional datasets, as well as by the absence of comparable causal learner modelling frameworks for the same problem. Extensive refinement and calibration of the learner model for our use case were beyond the objectives of this illustrative application; as a result, the proposed approach remains inferior in purely predictive terms, although not dramatically so ---a trade-off that is in any case a natural consequence of imposing interpretable causal assumptions and prioritising actionability, which are central objectives of the proposed approach. Finally,  the current software implementation of our method supports only exact BN inference, which may hinder its application to case studies involving many skills when the questions induce strong relations among them (App.~\ref{app:comp}). Supporting approximate BN inference is a necessary, and simple to achieve, future work.

Consequently, we do not claim evidence regarding the framework's impact on decision-making processes, or the correctness of the counterfactual estimates it produces, but articulate and exemplify a structural causal modelling approach to psychometric modelling to facilitate further methodological development of causal approaches in educational assessment. We identify broader validation through larger datasets or simulation studies, sensitivity analysis, and ultimately real-time deployment within an adaptive testing system, alongside addressing the scalability to larger skill sets, as a natural and necessary continuation of this line of research. 

\section*{Acknowledgments}
We thank Alberto Piatti for his involvement in the early discussions of this work. His thoughtful comments and encouragement contributed positively to the development of the ideas presented here.

This research was funded by the Swiss National Science Foundation (SNSF) under the National Research Program 77 (NRP-77) Digital Transformation (project number 407 740\_187246).


\appendix

\section{Inferential Complexity}\label{app:comp}

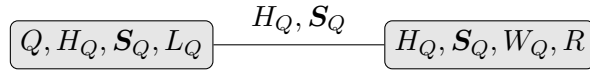
\begin{figure}[htp!]
\centering
\begin{tikzpicture}
\node[draw=black!100,fill=black!10,rounded corners=0.1cm] (1)  at (0,0) {$Q,H_Q,\bm{S}_Q,L_Q$};
\node[draw=black!100,fill=black!10,rounded corners=0.1cm] (3)  at (5,0) {$H_Q,\bm{S}_Q,W_Q,R$};
\draw[-]  (1) -- node[above] {$H_Q,\bm{S}_Q$}  (3);
\end{tikzpicture}
\caption{The two cliques of variables associated with $Q\in\bm{Q}$.}
\label{fig:tw}
\end{figure}

Let us now consider the computational complexity of the inferences performed within our protocol. Both the EM algorithm used to handle the latent nature of the exogenous variables and the inferences subsequently performed on the $n$ FSCMs it returns rely on BN inference over models whose underlying directed acyclic graph has a topology analogous to that in Fig.~\ref{fig:full}. Given a question $Q\in\bm{Q}$, let us denote as $\bm{S}_{Q} \subseteq \bm{S}$ the skills that are also parents of $Q$, i.e., the relevant skills to answering the question. Following, for instance, \cite{koller2009}, we \emph{moralise} the graph around the parents of $Q$ and $H_Q$ obtaining the cliques $(Q,H_Q,\bm{S}_Q,L_Q)$ and $(H_Q,\bm{S}_Q,W_Q,R)$. This yields, for each question $Q$, the induced graph shown in Fig.~\ref{fig:tw}. This is not a clique tree, for two reasons: (i) edges connect \emph{all} the cliques associated with the hints, since $R$ appears in every one of them; (ii) additional edges connect the cliques of different questions (and hints) whenever their relevant-skill sets have a non-empty intersection. A simple \emph{cutset} conditioning \citep{darwiche2009} on $R$ remove the edges responsible for (i). Moreover, since $L_Q$ and $W_Q$ each appear in only one clique, these variables can be handled by local computation. We further note that, in the inferences considered in Sect.~\ref{sec:inference}, both the questions and the hints are always either instantiated or queried.  The inferential complexity is therefore governed essentially by (ii), and depends on the extent to which the relevant-skill sets $\{\bm{S}_Q\}_{Q\in\bm{Q}}$ overlap across questions. In the worst case, this yields a \emph{treewidth} equal to $|\bm{S}|$, and the complexity is therefore exponential with respect to this value. Tighter bounds apply when distinct groups of questions draw on distinct groups of skills, since questions spanning multiple groups then induce only a sparser structure. Finally, we note that purely interventional queries require only direct inference in the FSCM, whereas counterfactual inferences require constructing a \emph{twin} network in which the endogenous variables are duplicated — modelling the factual and counterfactual worlds separately while sharing the exogenous variables. Since the endogenous variables involved in our counterfactual inferences are always either observed or queried, this duplication does not increase complexity. The discussion above concerns the complexity of exact inference at the level of a single FSCM; the bounds we derive for non-identifiable queries are inner approximations governed by the number of EM runs $n$. We refer the reader to \cite{zaffalon2023b} for a characterisation of the confidence levels with respect to $n$.

\section{Details on the Use-Case Assessment Protocol}\label{app:case}
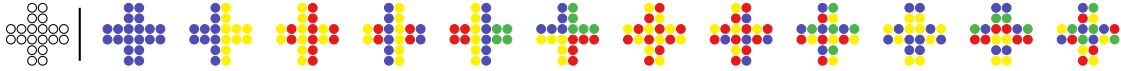
\begin{figure*}[!ht]
\centering
\begin{tikzpicture}[]
\tikzset{b/.style={draw,circle,scale=0.32,blue!40!gray,fill}}
\tikzset{g/.style={draw,circle,scale=0.32,yellow,fill}}
\tikzset{r/.style={draw,circle,scale=0.32,red!80!gray,fill}}
\tikzset{v/.style={draw,circle,scale=0.32,green!40!gray,fill}}
\tikzset{e/.style={draw,circle,scale=0.32,white,fill}}
\tikzset{w/.style={draw=black, circle, fill=white, scale=0.32}}
\tikzset{l10/.style={draw,rectangle,scale=0.42,black!99,fill}}   
\tikzset{l9/.style={draw,rectangle,scale=0.42,black!90,fill}}    
\tikzset{l8/.style={draw,rectangle,scale=0.42,black!81,fill}}    
\tikzset{l7/.style={draw,rectangle,scale=0.42,black!72,fill}}    
\tikzset{l6/.style={draw,rectangle,scale=0.42,black!63,fill}}    
\tikzset{l5/.style={draw,rectangle,scale=0.42,black!54,fill}}    
\tikzset{l4/.style={draw,rectangle,scale=0.42,black!45,fill}}    
\tikzset{l3/.style={draw,rectangle,scale=0.42,black!36,fill}}    
\tikzset{l2/.style={draw,rectangle,scale=0.42,black!27,fill}}    
\tikzset{l1/.style={draw,rectangle,scale=0.42,black!18,fill}}    
\tikzset{l0/.style={draw,rectangle,scale=0.42,black!9,fill}}     
\tikzset{n/.style={draw,rectangle,scale=0.42,black!0,fill}}
\matrix(m)[matrix of nodes,row sep=0.1mm,column sep=0.1mm]
{&&&&&|[w]|&|[w]|&&&&|[e]|&
 &&&&&|[b]|&|[b]|&&&&&
 &&&&&|[b]|&|[g]|&&&&&
 &&&&&|[g]|&|[r]|&&&&&
 &&&&&|[b]|&|[g]|&&&&&
 &&&&&|[g]|&|[b]|&&&&&
 &&&&&|[b]|&|[v]|&&&&&
 &&&&&|[g]|&|[r]|&&&&&
 &&&&&|[g]|&|[r]|&&&&&
 &&&&&|[b]|&|[v]|&&&&&
 &&&&&|[b]|&|[b]|&&&&&
 &&&&&|[b]|&|[b]|&&&&&
 &&&&&|[b]|&|[v]|&&&&\\
 &&&&&|[w]|&|[w]|&&&&&
 &&&&&|[b]|&|[b]|&&&&&
 &&&&&|[b]|&|[g]|&&&&&
 &&&&&|[g]|&|[r]|&&&&&
 &&&&&|[b]|&|[g]|&&&&&
 &&&&&|[g]|&|[b]|&&&&&
 &&&&&|[b]|&|[v]|&&&&&
 &&&&&|[r]|&|[g]|&&&&&
 &&&&&|[r]|&|[b]|&&&&&
 &&&&&|[r]|&|[g]|&&&&&
 &&&&&|[g]|&|[g]|&&&&&
 &&&&&|[v]|&|[v]|&&&&&
 &&&&&|[r]|&|[g]|&&&&\\
 &&&|[w]|&|[w]|&|[w]|&|[w]|&|[w]|&|[w]|&&&
 &&&|[b]|&|[b]|&|[b]|&|[b]|&|[b]|&|[b]|&&&
 &&&|[b]|&|[b]|&|[b]|&|[g]|&|[g]|&|[g]|&&&
 &&&|[g]|&|[r]|&|[g]|&|[r]|&|[g]|&|[r]|&&&
 &&&|[g]|&|[r]|&|[b]|&|[g]|&|[r]|&|[b]|&&&
 &&&|[r]|&|[r]|&|[g]|&|[b]|&|[v]|&|[v]|&&&
 &&&|[b]|&|[b]|&|[b]|&|[v]|&|[v]|&|[v]|&&&
 &&&|[g]|&|[r]|&|[g]|&|[r]|&|[g]|&|[r]|&&&
 &&&|[g]|&|[r]|&|[g]|&|[r]|&|[g]|&|[r]|&&&
 &&&|[b]|&|[v]|&|[b]|&|[v]|&|[b]|&|[v]|&&&
 &&&|[g]|&|[b]|&|[g]|&|[g]|&|[b]|&|[g]|&&&
 &&&|[v]|&|[b]|&|[r]|&|[r]|&|[b]|&|[v]|&&&
 &&&|[g]|&|[v]|&|[b]|&|[v]|&|[b]|&|[r]|&&\\
 &&&|[w]|&|[w]|&|[w]|&|[w]|&|[w]|&|[w]|&&&
 &&&|[b]|&|[b]|&|[b]|&|[b]|&|[b]|&|[b]|&&&
 &&&|[b]|&|[b]|&|[b]|&|[g]|&|[g]|&|[g]|&&&
 &&&|[g]|&|[r]|&|[g]|&|[r]|&|[g]|&|[r]|&&&
 &&&|[g]|&|[r]|&|[b]|&|[g]|&|[r]|&|[b]|&&&
 &&&|[r]|&|[r]|&|[g]|&|[b]|&|[v]|&|[v]|&&&
 &&&|[g]|&|[g]|&|[g]|&|[r]|&|[r]|&|[r]|&&&
 &&&|[r]|&|[g]|&|[r]|&|[g]|&|[r]|&|[g]|&&&
 &&&|[r]|&|[b]|&|[r]|&|[b]|&|[r]|&|[b]|&&&
 &&&|[r]|&|[g]|&|[r]|&|[g]|&|[r]|&|[g]|&&&
 &&&|[b]|&|[g]|&|[b]|&|[b]|&|[g]|&|[b]|&&&
 &&&|[r]|&|[r]|&|[g]|&|[g]|&|[r]|&|[r]|&&&
 &&&|[b]|&|[r]|&|[v]|&|[b]|&|[g]|&|[v]|&&\\
 &&&&&|[w]|&|[w]|&&&&&
 &&&&&|[b]|&|[b]|&&&&&
 &&&&&|[b]|&|[g]|&&&&&
 &&&&&|[g]|&|[r]|&&&&&
 &&&&&|[b]|&|[g]|&&&&&
 &&&&&|[g]|&|[b]|&&&&&
 &&&&&|[g]|&|[r]|&&&&&
 &&&&&|[g]|&|[r]|&&&&&
 &&&&&|[g]|&|[r]|&&&&&
 &&&&&|[b]|&|[v]|&&&&&
 &&&&&|[b]|&|[b]|&&&&&
 &&&&&|[b]|&|[b]|&&&&&
 &&&&&|[g]|&|[r]|&&&&\\
 &&&&&|[w]|&|[w]|&&&&|[e]|&
 &&&&&|[b]|&|[b]|&&&&&
 &&&&&|[b]|&|[g]|&&&&&
 &&&&&|[g]|&|[r]|&&&&&
 &&&&&|[b]|&|[g]|&&&&&
 &&&&&|[g]|&|[b]|&&&&&
 &&&&&|[g]|&|[r]|&&&&&
 &&&&&|[r]|&|[g]|&&&&&
 &&&&&|[r]|&|[b]|&&&&&
 &&&&&|[r]|&|[g]|&&&&&
 &&&&&|[g]|&|[g]|&&&&&
 &&&&&|[g]|&|[g]|&&&&&
 &&&&&|[r]|&|[g]|&&&&\\
&&&&&&&&&&&\phantom{}&&&&&&&&&&&\phantom{}&&&&&&&&&&&\phantom{}&&&&&&&&&&&\phantom{}&&&&&&&&&&&\phantom{}&&&&&&&&&&&\phantom{}&&&&&&&&&&&\phantom{}&&&&&&&&&&&\phantom{}&&&&&&&&&&&\phantom{}&&&&&&&&&&&\phantom{}&&&&&&&&&&&\phantom{}&&&&&&&&&&&\phantom{}&&&&&&&&&&&\phantom{}\\
 };
\draw[thick] (m-1-11.east) -- (m-6-11.east);
\end{tikzpicture}
\caption{Blank cross array (left) and the 12 CAT schemes.}
\label{fig:CATschemes}
\end{figure*}

The CAT is a battery of unplugged tasks designed to assess the algorithmic skills component of computational thinking in pupils aged from 3 to 16 years. 
In the CAT assessment, students are asked to verbally instruct a tutor to reproduce, on a blank cross array, the 12 target arrays in Fig.~\ref{fig:CATschemes}, consisting of 20 coloured circles each. A barrier prevents the student from seeing how the tutor is colouring. Students have access to two forms of assistance: they can (i) request a blank cross array to support their verbal instructions by pointing to the circles to colour; (ii) remove the barrier and obtain visual feedback of the result of their instructions. In \cite{piatti2022ct}, each set of student instructions is referred to as an algorithm and categorised into three levels: 0D (zero-dimensional), where students specify colours circle by circle; 1D, which incorporates structures like rows and columns; 2D, which includes loops. Student autonomy is also evaluated based on the type of assistance they require to complete a task. The lowest level of competence is when the student requires both the blank cross array and visual feedback; the intermediate level involves using only the blank cross array; the highest using none.

\end{document}